\documentclass{sig-alternate}

\usepackage{amsmath}
\usepackage[english]{babel}
\usepackage[shrink=25]{microtype}
\usepackage{moreverb}
\usepackage{nicefrac}
\usepackage{graphicx}
\usepackage{hyperref}
  \hypersetup{
    colorlinks=true,
    linkcolor=black,
    citecolor=black,
    menucolor=black,
    urlcolor=black,
    breaklinks=true,
    pageanchor=true,
    pdfauthor={Juergen Mueller and Gerd Stumme},
    pdftitle={Gender Inference using Statistical Name Characteristics in Twitter},
    pdfdisplaydoctitle=true,
    pdflang={english}
  }

\usepackage[utf8]{inputenc}
  \usepackage{tipa} 
  \DeclareUnicodeCharacter{0252}{\textturnscripta}
  \DeclareUnicodeCharacter{028C}{\textturnv}
  \DeclareUnicodeCharacter{03B5}{$\varepsilon$}
\usepackage[numbers,sort&compress]{natbib}
\usepackage{url}
\usepackage{xspace}

\newcommand{\fig}[3]
{
  \begin{figure}[#1]
    \centering
    \includegraphics[width=\columnwidth]{fig/#2}
    \caption{#3}
    \label{fig:#2}
  \end{figure}
}
\newcommand{\cfig}[4]
{
  \begin{figure}[#1]
    \centering
    \includegraphics[width=#2\columnwidth]{fig/#3}
    \caption{#4}
    \label{fig:#3}
  \end{figure}
}

\newcommand{\fsource}[1]{\footnote{Source: #1}}
\newcommand{\furl}[1]{\fsource{\url{#1}}}

\newcommand{\censusTotal}{5,163\xspace}
\newcommand{\censusUnisex}{331\xspace}

\newcommand{\namDictTotal}{45,513\xspace}

\newcommand{\sampleBinSize}{3,471\xspace}

\newcommand{\tweetsFemale}{5,589\xspace}
\newcommand{\tweetsMale}{3,471\xspace}
\newcommand{\tweetsNullCensus}{4,790\xspace}

\newcommand{\tweetsNullDiff}{9.7~\%\xspace}
\newcommand{\tweetsNullNamDict}{4,323\xspace}
\newcommand{\tweetsNullNamDictShare}{47.7~\%\xspace}
\newcommand{\tweetsNullSvm}{288\xspace}
\newcommand{\tweetsNullSvmShare}{3.2~\%\xspace}
\newcommand{\tweetsTotal}{9,060\xspace}
\newcommand{\preprocessingChange}{426\xspace}
\newcommand{\preprocessingChangeShare}{8.89\xspace}

\providecommand{\tightlist}{\setlength{\itemsep}{0pt}\setlength{\parskip}{0pt}}

\newcommand{\figref}[1]{Figure~\ref{fig:#1}}

\newcommand{\tabref}[1]{Table~\ref{tab:#1}}
\newcommand{\secref}[1]{Section~\ref{sec:#1}}

\newcommand{\namdict}{nam\_dict\xspace}
\newcommand{\census}{Census\xspace}
\newcommand{\namchar}{NamChar\xspace}

\begin{document}

\CopyrightYear{2016}
\setcopyright{acmlicensed}
\conferenceinfo{MISNC, SI, DS '16,}{August 15 - 17, 2016, Union, NJ, USA}
\isbn{978-1-4503-4129-5/16/08}\acmPrice{\$15.00}
\doi{http://dx.doi.org/10.1145/2955129.2955182}

\title{Gender Inference using Statistical Name Characteristics in Twitter}
\numberofauthors{2}
\author{
	\alignauthor
	Juergen Mueller\\
	\affaddr{University of Kassel}\\
	\affaddr{Research Center for Information System Design (ITeG)}\\
	\affaddr{Pfannkuchstr. 1}\\
	\affaddr{34121 Kassel, Germany}\\
	\email{mueller@cs.uni-kassel.de}
	\alignauthor
	Gerd Stumme\\
	\affaddr{University of Kassel}\\
	\affaddr{Research Center for Information System Design (ITeG)}\\
	\affaddr{Pfannkuchstr. 1}\\
	\affaddr{34121 Kassel, Germany}\\
	\email{stumme@cs.uni-kassel.de}
}
\maketitle

\begin{abstract}

Much attention has been given to the task of gender inference of Twitter users. Although names are strong gender indicators, the names of Twitter users are rarely used as a feature; probably due to the high number of ill-formed names, which cannot be found in any name dictionary. Instead of relying solely on a name database, we propose a novel name classifier. Our approach extracts characteristics from the user names and uses those in order to assign the names to a gender. This enables us to classify international first names as well as ill-formed names.

\end{abstract}





\section{Introduction}
\label{sec:introduction}
Both academia and companies have genuine interest in understanding the gender distribution in social networks. Social scientists could study gender as an influence on human behavior in online communities~\citep{hong2015belief} or on scientific and technological productivity from countries~\citep{frietsch2009gender}. The industry would gain additional information about their customers, which allows them to improve the efficiency of targeted advertisements~\citep{berney-reddish2006sex}.

Twitter is currently one of the biggest, most important, and scientifically best covered social networks. Data are mostly public and it is well understood by academia, which allows good comparisons in return. Unfortunately, explicit gender information is not included in the Twitter data. Therefore, it has to be inferred, for example, using machine learning methods. Accordingly, automated gender inference of Twitter users is a relevant topic of research. Amongst those, classification using Support Vector Machines~(SVMs) have been found to be the best gender inference approaches \citep{burger2011discriminating,liu2013whats,pennacchiotti2011machine}. Examples for the most commonly used features are bag of words, n-grams, hashtags, or the friend-follower ratio.

Surprisingly, the self-reported names\footnote{Twitter has two names per user. We refer here to the ``real name'' that is displayed on the profile page. There is also the ``username'' that is used as unique identifier for every user.} of the users are rarely used as a classification feature. This was mentioned by \citet{liu2013whats} who conducted some experiments that are based on the self-reported name of the Twitter users as additional feature.

One could argue that the self-reported names and profile images offer no guarantee to be true. But according to \citet{herring2014gender}, the number of users who masquerade themselves for someone else is not statistically significant and can be ignored in most cases.

Making use of some name dictionary seems to be an obvious solution to get gender information about a given name. Such dictionaries contain known names that are actually used by human beings. Twitter users, however, did not restrict their choice to this set of names nor does Twitter enforce it. Users are named with their actual names, made-up names, or nicknames.

We will present a new classifier named \namchar that assigns gender labels to first names based on their written form. It uses methods from the study of onomastics to extract characteristics from the name that correlate with the two genders. Our paper tries to answer the following research questions:

\begin{enumerate}
  \tightlist
  \item\label{q:broader-dictionary} In their outlook, \citet{liu2013whats} considered a bigger name dictionary as promising improvement for their gender score. Therefore, we answer the question ``does a broader name dictionary improve the performance of the gender score?''
  \item\label{q:namchar} We want to improve the result of \citeauthor{liu2013whats} by using name characteristics. This enables the classification of names that cannot be found in a name dictionary. This leads to the following two sub-questions:
  \begin{enumerate}
    \tightlist
    \item\label{q:namchar:score} ``Are name characteristics able to improve the performance of the gender score?''
    \item\label{q:namchar:overall} ``Are name characteristics able to improve the overall performance of \citeauthor{liu2013whats}'s gender inference method?''
  \end{enumerate}
\end{enumerate}

In the sequel, we always use the word ``gender'' as synonym for ``gender of first names'' (with the two instances ``female'' and ``male'') unless we explicitly talk about the gender of people. We could add ``unisex'' as a third gender, but have refrained from it, because we have no ground truth covering ``unisex'' names on Twitter.


\section{Related Work}
\label{sec:related-work}

A gender-labeled dataset of names can be very beneficial for gender classification. However, it cannot be used as sole data source, because Twitter names are not limited to real names. There are many sources for name lists on the Internet. Most sources, however, give no information about the quality of the data. Nevertheless, we found four data sources of trustworthy quality:

\begin{itemize}
  \tightlist
  \item The list of most frequently chosen baby names for every year since 1880 as published by the US Social Security Agency.\fsource{File ``names.zip'' from \url{https://ssa.gov/oact/babynames/limits.html}}
  \item The names of the participants of the 1990 \census as published by the US Census Bureau.\fsource{File ``dist.female.first'' and ``dist.male.first'' from \url{http://census.gov/topics/population/genealogy/data/1990_census/1990_census_namefiles.html}\label{footnote:census}}
  \item Jörg Michael published a self collected dataset of names.\fsource{File ``nam\_dict.txt'' from \url{https://heise.de/ct}, soft-link 0717182 (c) by Jörg MICHAEL, Hannover, Germany, 2007-2008\label{footnote:namdict}}
  \item Wikipedia contains dedicated pages covering first names that are available as download from Wikimedia.\furl{https://dumps.wikimedia.org}
\end{itemize}

\citet{tang2011whats} proposed a gender inference classifier for Facebook users from New York City. They collected 1.67~million profiles and extracted a gender-labeled list of names from this dataset. They used the top baby names published by the US Social Security Agency, their list of collected Facebook names, data from the Facebook fields ``relationship status'', ``interested in'', and ``looking for'', as well as information about the Facebook friends to predict the gender of the users with an accuracy of up to 95.2~\%. The authors decided to randomly assign a gender if a user's name is not found in their list of names, which leaves room for further improvement.

\citet{karimi2016inferring} compared five gender inference tools in the realm of research names. They used an image-based gender inference on the five first search engine results using the first and last name. Their approach, however, works with actual names to query the search engine. Twitter names mostly do not fall into this category.

\citet{liu2013whats} proposed a novel gender inference classifier using an SVM that uses the names of Twitter users as additional feature. As Twitter offers no gender information and there was no publicly available labeled dataset at the time, they created a classified dataset on their own. They used Amazon Mechanical Turk (AMT) workers for determining the gender of users based on their profile images. \citeauthor{liu2013whats} introduced a gender-name association score (from now on referred to as ``gender score'') for all names in the \census data. The gender score is computed using the gender distribution of each name and reflects how often it has been given to a male or female person. They used common features for their classification and added the gender score as additional feature. Their results show that the use of this gender score reduced the classification error rate by 11.4~\%, which they improved further to 22.8~\% by the use of a threshold value. Their approach, however, can only work to its full potential, when the name of concern is represented in the \census data. But Twitter is an international network with users from all around the world. Accordingly, they could not assign a gender score to about two-thirds of the users. Even a hypothetical dataset with gender-labeled names covering all countries of the world could not classify all Twitter users, because Twitter does no force its users to use real names.

One attempt to infer the gender of Twitter users is the analysis of the self-reported name directly. \citet{slater1985gender} and \citet{cutler1990elizabeth} discovered a significant correlation between name characteristics and the gender of North American names. Their findings were later transferred successfully to German names by \citet{oelkers2003naming} and \citet{seibicke2008die}. The English and German findings can be used to identify the gender of a given name based on the number of syllables, number of vowels, number of consonants, vowel brightness, and ending character. Among those, the ending sound is the strongest. The majority of female names ends with a vowel, while most names that end with a consonant are male. This implication is true for about 60~\% of North American and 80~\% of German cases. Unfortunately, some characteristics depend on the pronunciation, which is problematic, because we have no information about the origin of the names and their actual pronunciation.

\cfig{tb}{0.3}{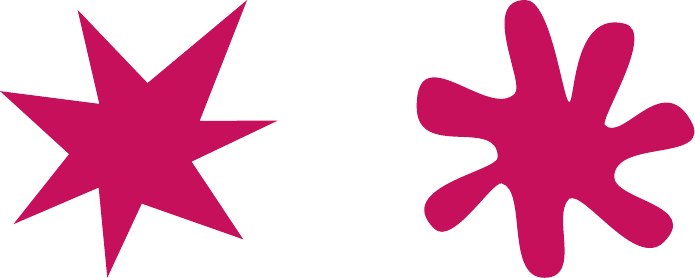}{Shapes that are used to analyze the perception of speech sounds and shapes. The left shape is referred to as ``Kiki'' and the right one as ``Bouba''.}

Another promising approach to extract the gender of a name from its written word was discovered by \citet{sidhu2015whats}. They analyzed the use of the Bouba/Kiki effect on given names. The Bouba/Kiki effect describes a non-arbitrary mapping between a speech sound and the visual shape of objects (see \figref{bouba-kiki}) \citep{koehler1970gestalt}. \citeauthor{sidhu2015whats} asked the participants of their study to assign a given English first name to two shapes (i.e., Bouba or Kiki) that are shown to them during the experiment. Every name was assigned to either Bouba or Kiki beforehand based on the findings of previous research \citep{maurer2006shape,nielsen2011sound}. \citeauthor{sidhu2015whats} found a relationship between Bouba with female names and Kiki with male names through the answers of the participants. These characteristics, however, suffer from the same challenge as parts of the aforementioned name characteristics: we do not know the actual pronunciation. Further, these findings only account for a correlation in the English language.


\section{Datasets}
\label{sec:datasets}
This section describes the data used throughout this paper. First we present the gender-labeled name data that can be used to get the true gender from a given name. Second we describe the data that is used for our Twitter experiments. Note that we considered only ``male'' and ``female'' as potential genders of people, ignoring sexual identities like ``Cisgender'' or ``Transgender'', because there is no available ground truth for that.

\subsection{Gender-labeled Name Data}
\label{sec:name-data}
A reliable gender-labeled name dataset is required in order to make assumptions about the performance of a name-driven gender inference classifier. We decided to use the Census\textsuperscript{\ref{footnote:census}} and the \namdict data\textsuperscript{\ref{footnote:namdict}} throughout our experiments for the following reasons: First, the \census data was used by \citet{liu2013whats} and we will use their results as baseline in \secref{gender-inference}. Second, the \namdict dataset is the biggest dataset we know of that covers non-American names as well.
Following is a short description of both datasets:

\textbf{\census}: The \census dataset was created in 1990 by the US Census Bureau and consists of three files, one for female first names, one for male first names, and one for all last names. Each file contains for each upper-cased name, the frequency of its usage in percent, the cumulative usage frequency in percent, and its rank. The dataset contains only names that are used by at least 0.001~\% of US citizens, which results in \censusTotal names and corresponds to about 90~\% of the population of the United States of America. We use only the files for first names and therein only the names and their frequencies.

\textbf{\namdict}: The name dictionary \namdict in version~1.2 was collected by Jörg Michael in 2008. It is with \namDictTotal first names and nicknames more than eight times bigger than the \census data. The name dictionary contains names covering 51 geographic regions of the world. All names are stored in their actual writing (i.e., no upper- or lower-casing was applied) alongside information about their gender association and how often they are used in every geographic region. The gender association is encoded into seven categories, which provides the following gender information: ``mostly male/female name'', ``female/male name if first part of the name'', and ``male/female/unisex first name''.

\tabref{name-data-statistics} shows the different gender labels and numbers of names assigned to them in both datasets. Note that the \census data does not contain dedicated gender labels. The names are assigned to a gender using their presence in one of two files. Therefore, the number of unisex names is not explicitly given. After intersecting the names from both data files, we found that \censusUnisex names in the \census data are used by both genders. The names from the \namdict dataset are stored and labeled in a single file. Their labels directly correspond to those categories used in \tabref{name-data-statistics}.
The \census data contain a much bigger share of female names than the \namdict data. The \namdict dataset contains far more unisex names.

\begin{table}[tb]
\centering
\caption{Number of names contained in each gender-labeled name dataset. The numbers show the labels as they are retrieved from the dataset files.}
\label{tab:name-data-statistics}
\begin{tabular}{lrr}
  \hline
Gender & Census & nam\_dict \\
  \hline
Male first name & 888 & 18,204 \\
  Mostly male name & -- & 915 \\
  Male name if first part & -- & 7 \\
  Female first name & 3,944 & 17,328 \\
  Mostly female name & -- & 722 \\
  Female name if first part & -- & 8 \\
  Unisex first name & 331 & 8,329 \\
   \hline
Total number of names & 5,163 & 45,513 \\
   \hline
\end{tabular}
\end{table}

\subsection{Twitter Data}
\label{sec:data:twitter}
Next, we describe the data that will be used for the gender inference of Twitter users. It consists of two parts, namely a gender-labeled reference and a corresponding crawled Twitter sample. Following is a short description of both datasets:

\textbf{Ground truth}: Reliable information about the actual gender of the Twitter users is essential in order to conduct gender inference experiments. Such data is hard to find and rarely shared. \citet{liu2013whats} were faced with the same issue and, therefore, created their own dataset and made it publicly available.\fsource{File ``label.json'' from \url{http://networkdynamics.org/static/datasets/LiuRuthsMicrotext.zip}} They created this dataset by randomly collecting 50,000 users from Twitter gardenhose including their history of Tweets and profile information. The only condition was that those users had posted at least 1,000 Tweets. \citeauthor{liu2013whats} presented the profile images of those Twitter users to Amazon Mechanical Turk (AMT) workers, who had the task to label these pictures as ``male'', ``female'', or ``unknown''. Every profile image was given to three AMT workers and got a gender label only if all three of them agreed on the gender. This procedure did not only ensure a reliable classification into ``male'' and ``female'', but also made it possible to remove cooperate and other non-personal accounts with logos in their profile images by using the label ``unknown''. The resulting dataset consisted of 12,681 Twitter user identifier with assigned gender labels (4,449 male and 8,232 female).

\textbf{Twitter sample}: We collected all messages and relevant profile information from the Twitter users contained in the ground truth data created by \citeauthor{liu2013whats}. Only \tweetsTotal of the 12,681 users could still be accessed via the Twitter API to date. This breaks down to \tweetsMale male and \tweetsFemale female users (according to the AMT workers). Reasons for the decreased amount of accessible users could be that the dataset was collected before 2013 and many users deactivated their account or increased their privacy level since then. \citeauthor{liu2013whats} had not used all 12,681 users either, but made a sub-sample of 4,000 male and 4,000 female users to create equal sized data for both genders. However, we could only access \sampleBinSize profiles from male users. Following the equal-sized approach of \citeauthor{liu2013whats}, we created a sub-sample of \sampleBinSize users per gender.


\section{NamChar}
\label{sec:name-characteristics}
Our first contribution is a name classifier that works on the written word of a first name. We first present the task description, following a discussion of the possible name characteristics that could be used. At the end of this section, we present our novel name classifier \namchar and its configuration.

\subsection{Task Description}
\label{sec:gender-inference:task-description}
Given is a list of first names with their gender information. We want to find a classifier that is able to assign gender labels to those names, using only the written name itself.

\subsection{Data Preparation}
We choose to use the \namdict dataset because it contains the most extensive name information at hand. As shown in \tabref{name-data-statistics}, the dataset contains many more categories and, therefore, had to be pre-processed as follows: The gender information was transformed by assigning the categories ``male first name'', ``mostly male name'', and ``male name if first part of the name'' to ``male'', the categories ``female first name'', ``mostly female name'', and ``female name if first part of the name'' to ``female'', and the category ``unisex first name'' was removed, because it could decrease the accuracy.

\subsection{Feature Selection}
First, we need to identify a set of characteristics that can be extracted from the written word of a name. They should be easy to extract while giving enough information about the corresponding gender. Therefore, we will first discuss the characteristics that were identified in onomastic research \citep{slater1985gender,cutler1990elizabeth,oelkers2003naming,seibicke2008die}:

\begin{itemize}
  \tightlist
  \item Number of syllables: Female names tend to have more syllables than their male counterparts.
  \item Number of consonants: Male names tend to contain more consonants than female names.
  \item Number of vowels: Female first name tend to contain more vowels than male names.
  \item Vowel brightness: Female names contain more brightly emphasized vowels than male names.
  \item Ending character: Female names end more often with a vowel while male names tend to end with a consonant.
\end{itemize}

Further, one could use the findings of \citet{sidhu2015whats} and count vowels and consonants that are associated with Bouba or Kiki. \citeauthor{sidhu2015whats} discovered a relationship of Bouba with female first names and Kiki with male first names. Therefore, one can use four additional name characteristics using the encoding schema for Bouba/Kiki that is provided by previous work as follows \citep{maurer2006shape,nielsen2011sound}:

\begin{itemize}
  \tightlist
  \item Number of Bouba consonants: Female name can be identified by counting the voiced consonants ``b'', ``l'', ``m'', and ``n''.
  \item Number of Bouba vowels: Female name can be identified by counting the rounded vowels ``u'', ``o'', and ``ɒ''.
  \item Number of Kiki consonants: Male name can be identified by counting the voiceless stop consonants ``k'', ``p'', and ``t''.
  \item Number of Kiki vowels: Male name can be identified by counting the unrounded vowels ``i'', ``e'', ``ε'', and ``ʌ''.
\end{itemize}

All of these characteristics can be extracted from a written name. The ending character is the strongest amongst those characteristics as the majority of female names end with a vowel, while most names that end with a consonant are male. However, there are some limitations to the use of some of those characteristics. First do all characteristics depend on the pronunciation of the name, except of the number of consonants, number of vowels, and the ending character. Unfortunately, there is no indication about the pronunciation of a name in Twitter. Second, the Bouba/Kiki findings have not been transferred to other languages before and can, therefore, only ensure for a correlation in the English language. Hence, we should be very sensitive while using these characteristics!

We conducted a logistic regression in order to identify the strong predictors amongst the name characteristics. These predictors later will be used as variables in our classifier. During our experiments, all pronunciation-dependent variables where computed using the English language.

\begin{table}[tb]
\centering
\caption{Summary of the interval scaled variables.}
\label{tab:variable-summary}
\begin{tabular}{lrrrr}
  \hline
Name & Min & Max & Mean & SD \\
  \hline
number of consonants & 0 & 10 & 3.544 & 1.284 \\
  number of vowels & 0 & 9 & 2.847 & 0.915 \\
  number of syllables & 1 & 7 & 1.754 & 0.728 \\
  number of bouba consonants & 0 & 4 & 1.054 & 0.766 \\
  number of bouba vowels & 0 & 4 & 1.620 & 0.681 \\
  number of kiki consonants & 0 & 3 & 0.241 & 0.464 \\
  number of kiki vowels & 0 & 1 & 0.073 & 0.260 \\
   \hline
\end{tabular}
\end{table}

The data consists of the gender and the characteristics for every non-unisex name in the \namdict dataset mentioned above. The data consists of 18,204 (51.23 \%) male and 17,328 (48.77 \%) were female names. The research hypothesis posed to the data is that ``the likelihood that a name is used by women is related to its characteristics''. Thus, the outcome variable gender is a name being used by female (1 = yes, 0 = no). The ending character is coded as 1 = vowel and 0 = consonant. The distribution was 58 \% (n = 21,729) vowels and 42 \% (n = 15,455) consonants. The description of the interval scale-based variables is shown in \tabref{variable-summary}. We observed a variance inflation factor~(VIF) of 10.1 for the vowel brightness after the first run. Following the rules given by \citet{kutner2005applied}, we removed this variable in order to remove the multicollinearity.

\begin{table}[tb]
\centering
\caption{The observed and predicted frequencies for gender by logistic regression. Sensitivity = 66.39~\%. Specificity = 76.50~\%. False positive = 20.11~\%. False negative = 38.18~\%.}
\label{tab:namchar-logistic-regression-predicitve-power}
\begin{tabular}{lrrr}
  \hline
  & \multicolumn{2}{c}{Predicted} & \\
Observed & female & male & Correct \% \\
  \hline
female & 14,426 & 7,303 & 66.39 \\
  male & 3,632 & 11,823 & 76.50 \\
  Overall \% correct &  &  & 70.59 \\
   \hline
\end{tabular}
\end{table}

\begin{table*}[t]
\centering
\caption{Logistic regression analysis of 37,184 name's gender (1 = female, 0 = male) by R version 3.2.4 (2016-03-10) using the glm function from the nlme package version 3.1-126. Significance codes: $<.01$ = `**', $<.05$ = `*'. Nagelkerke's $R^2$ = .2425.}
\label{tab:namchar-logistic-regression}
\begin{tabular}{lrrrrrlr}
  \hline
& $\beta$ & SE $\beta$ & Wald's $\chi^2$ & $df$ & $p$ &  & $e^\beta$ \\
  \hline
Number of consonants & -0.0089 & 0.0118 & 0.5680 & 1 & .4510 &  & 0.9912 \\
  Number of vowels &  0.0668 & 0.0176 & 14.4286 & 1 & .0001 & ** & 1.0691 \\
  Number of syllables &  0.0243 & 0.0179 & 1.8441 & 1 & .1745 &  & 1.0246 \\
  Ending character  (1 = vowel, 0 = consonant) &  1.8976 & 0.0300 & 3,998.1399 & 1 & .0000 & ** & 6.6701 \\
  Number of bouba consonants &  0.2911 & 0.0162 & 320.9642 & 1 & .0000 & ** & 1.3379 \\
  Number of bouba vowels &  0.0779 & 0.0239 & 10.6099 & 1 & .0011 & ** & 1.0810 \\
  Number of kiki consonants & -0.0598 & 0.0248 & 5.7863 & 1 & .0162 & * & 0.9420 \\
  Number of kiki vowels &  0.0898 & 0.0442 & 4.1275 & 1 & .0422 & * & 1.0939 \\
  Constant & -1.8224 & 0.0502 & 1,316.4648 & 1 & .0000 & ** & NA \\
   \hline
\end{tabular}
\end{table*}

The results of the final model are shown in \tabref{namchar-logistic-regression}. The number of consonants and syllables are not significant ($p=.4510$, respectively $p=.1745$). The number of Kiki vowels and consonants are significant on a $<.05$ level and the remaining variables are all significant on a $<.01$ level. The validity of the predicted probabilities is documented in \tabref{namchar-logistic-regression-predicitve-power}. According to the table is \namchar more likely to predict male names correctly than female---66.39~\% for female and 76.50~\% for male names---while both stay well above random. The overall prediction was 70.59~\%, an improvement over the chance level.

\subsection{Classification Model}
Inspired by \citeauthor{liu2013whats}, we applied an SVM (i.e., LIBSVM \citep{chang2011libsvm}) with the radial basis function as kernel. Following the logistic regression results in \tabref{namchar-logistic-regression}, we include all variables except of the vowel brightness and the number of consonants and syllables to our SVM model. We used a grid search technique with a 10-fold cross-validation with three repeats to find the best parameter that are $\gamma$ = 0.1745521 and Costs = 1 with an accuracy of 70.9~\% and a Kappa value of 0.419.

The default approach to assign genders to the names without a classification would be through random guessing. On the other hand, our \namchar is able to classify a name correctly in 70.9~\% cases as mentioned in the previous paragraph. This is a considerable improvement compared to random guessing. We expect that the model works as well on unknown names.

We will use this approach in the following section where it will be used to classify Twitter names. Most Twitter names are not in any gender-labeled name database. Therefore, \namchar will be used when a name that is not represented in the database of known names needs to be classified.


\section{Twitter Gender Inference}
\label{sec:gender-inference}
The second contribution of this paper is to show how gender inference for Twitter users can be improved by applying \namchar. We recall the gender score of \citet{liu2013whats} and introduce our refinement, the \namchar score. Then we show how the gender classification of Twitter users can be improved by extending the \namchar Threshold Classifier. The key idea of both approaches is to look up the name in a database. The stored score is used if the database contains the name. If this is not the case, \citet{liu2013whats} use a score of 0.0 whereas our approach applies \namchar.

It is important to note that the \namdict data is used here as input data, rather than as ground truth as in \secref{name-characteristics}.

\subsection{Gender Scores}
\label{sec:gnas}
The gender score $s(n)$ as introduced by \citet{liu2013whats} reflects the information that some names are more frequently assigned to male than to female people. For example, the names ``John'' ($s(n)=0.993$) and ``Ashley'' ($s(n)=-0.912$) are clearly associated to one gender while ``Berry'' ($s(n)=0.714$) and ``Kim'' ($s(n)=-0.728$) are used by both genders. The gender score on the \census data is computed using the following equation: $s(n) = \frac{M(n) - F(n)}{M(n) + F(n)}$, where $n$ is the name of interest, $M(n)$ is the number of times $n$ is assigned to men, and $F(n)$ is the number of times $n$ is assigned to women. The score ranges from $-1.0$ for names that are given solely to women to $1.0$ for names that are given to men only. For names that are not present in the \census data, the score is set to $0.0$.

\fig{t}{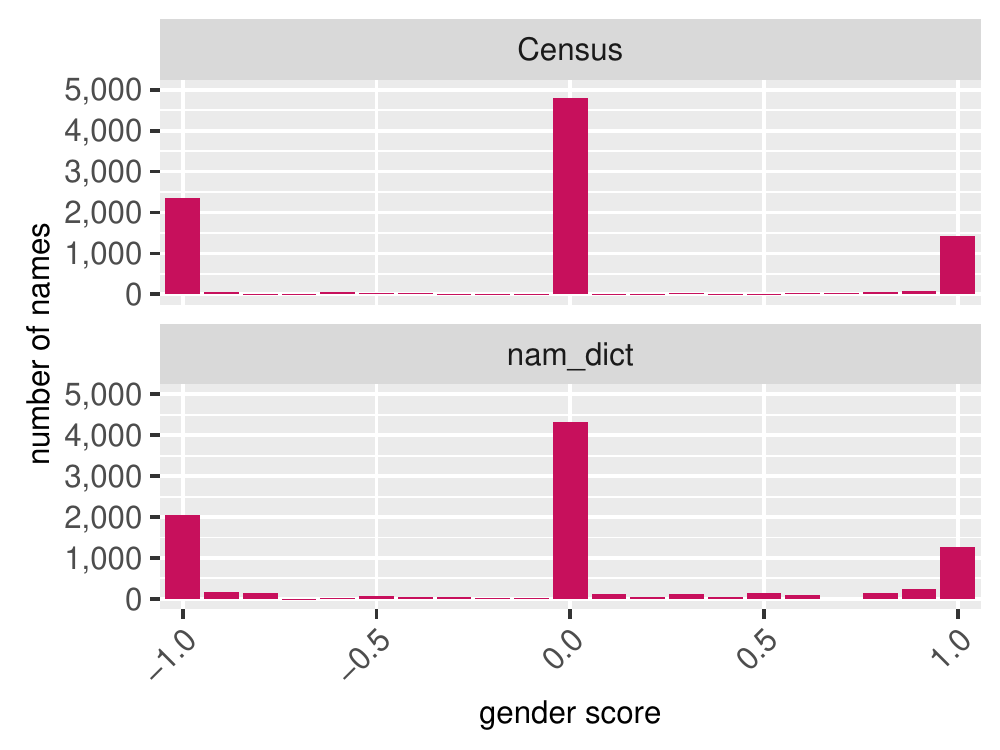}{The gender score of all studied Twitter users based on the \census and \namdict data, respectively. The $x$-axis shows the binned score values (from $-1.0$ for ``female'' to $1.0$ for ``male'') while the $y$-axis represents the number of names with that score.}

The gender score can be calculated directly from the \census data, because they contain explicit frequencies for the usage of the names by each gender. The \namdict data lacks such information, because it contains only binned information about the usage. In order to adapt the gender score to the \namdict data, we will translate the categories ``male/female first name'' to $\pm1.0$ and ``mostly male/female name'' to a value of $\pm0.8$. ``female/male name if first part of the name'' will be used as the description suggests; that is $\pm1.0$ if it actually is the first part of a name and $0.0$ otherwise. The \namdict data may contain multiple entries for some names, if they are used for different genders in different geographic regions. We use the average of the gender scores of all occurrences in those cases.

\figref{gender-score-distributions} shows the distribution of all names of the Twitter dataset by their gender score. Both plots show the same tri-modal characteristics as in the original work by \citet{liu2013whats}. Most classified names are unambiguously assigned to one of the two genders, but of particular note is that the vast majority of names has a score of $0.0$. These names cannot be matched to an entry of the name database and can therefore not be used in Liu and Ruth's algorithm. Amongst those are uncommon names like ``Wela P Msimanga'', ``Jessele Competente'', or ``Laketa Page'' that sound like actual names, but are not included in either of our name databases. \citet{liu2013whats} already identified this issue and separated the unknown names into five categories:\footnote{Examples adopted from \citet{liu2013whats}.}

\begin{enumerate}
  \tightlist
  \item Unknown names like ``Lim'' (which is actually a unisex name according to the \namdict data). Those names are not present in the \census data. Therefore, \citeauthor{liu2013whats} were not able to assign a meaningful gender score to those names. This category is largely covered by the use of the \namdict data.
  \item Nicknames and name abbreviations like ``Big Daddy C'' or ``CJ Sullivan''. Those names do not match to existing names. Nicknames, however, still contain some name characteristics. These signals will be exploited by our \namchar classifier.
  \item Mangled names like ``AlanLeong'' or ``[-!Raphael-]''. Those names contain valid given names, but are run together or decorated. Decorated names are less of a problem during our experiments, because we remove all decorating symbols from the name string. Names that are run together are more difficult because we lose the ending character as a feature.
  \item Usernames like ``swagboiboo504''. Those names read like user names or mail addresses. They could contain some gender characteristics similar to nicknames or mangled names, because they are likely to either contain them.
  \item Non-names like ``Married To Bieber'', ``Apple Support'' or ``The Great Gatsby''. Those names are no first names in any sense and are not intended to be so. Therefore, it is very unlikely that we could use them in a meaningful way. Worse, those names will likely reduce the accuracy of our approach.
\end{enumerate}

Using a larger database of first names (i.e., the \namdict data) increased the number of assigned gender scores, but the number of names with a gender score of 0.0 still remains very high. While \tweetsNullCensus user names received ay value of $0.0$ with the \census data, \tweetsNullNamDict did so with the \namdict data; a reduction of \tweetsNullDiff.  The fact that still \tweetsNullNamDictShare of our Twitter users do not use common names is a strong signal that relying solely on name lists can only be part of a gender inference solution. Based on our observations in \secref{name-characteristics}, we expect to assign helpful gender scores to at least some of those \tweetsNullNamDict users using \namchar.

We introduce the \textit{\namchar gender score} that addresses this issue. It applies \namchar on those names that could not be found in the name dataset and uses the classification probability as gender score.
This reduces the number of users without a meaningful gender score to \tweetsNullSvm, which is only \tweetsNullSvmShare of all users. Those remaining names are without exception names that consist solely of Unicode characters that could not be translated to common Latin letters, like Arabic names.

\begin{table}[tb]
\centering
\caption{Evaluation results for the gender score on both datasets and the NamChar score in the Census data showing the resulting AUC, accuracy~(ACC), recall~(REC), and Kappa~(K) values.}
\label{tab:gender-score-dictionary-comparisson}
\begin{tabular}{llrrrr}
  \hline
Score & Dataset & AUC & ACC & REC & K \\
  \hline
Gender score & Census & 0.786 & 0.628 & 0.429 & 0.324 \\
  Gender score & nam\_dict & 0.782 & 0.621 & 0.420 & 0.312 \\
  NamChar score & Census & 0.804 & 0.692 & 0.630 & 0.394 \\
   \hline
\end{tabular}
\end{table}

Next, we compare both gender scores as predictor for the true genders of all collected Twitter users (N = 9,060) in order to find out whether or not a broader name dictionary improves the gender score (Research Question~\ref{q:broader-dictionary}). The first two rows of \tabref{gender-score-dictionary-comparisson} show the results of our experiments. The two datasets lead to similar results. A comparison of both datasets shows comparable results across all evaluation measures: For a paired-samples $t$-test (using SPSS version~23) we scaled the gender scores to the range of 0 to 1---using $\nicefrac{s(n)}{x}+5$ for male and $\nicefrac{-s(n)}{x}+5$ for female names. The test shows that there is a slightly, but significant difference between both results, $t(9,059) = 6.10, p<.001$, in which the gender scores based on the \census data obtained better results (\census: M=0.683, SD=0.277; \namdict: M=0.669 SD=0.282). This disproves the assumption of \citeauthor{liu2013whats} that a broader name database could improve their approach.

\citet{liu2013whats} where not able to assign a valid gender score to two thirds of the Twitter users. \tweetsNullDiff more users receive a gender score if the \census data are replaced by the \namdict data, but on the expense of worse results. Therefore, we continue to use the \census data, but will present in the sequel a better approach to assign gender scores to those names that are not contained in the dataset.

We address these cases by using our \namchar score, which applies \namchar on all instances where the gender score equaled zero. The corresponding results can be found in Row~3 of \tabref{gender-score-dictionary-comparisson}. The overall results are better compared to the original approach (Row~1) for all measures. The biggest improvement is to be found for Recall which increases by 44~\% from 0.429 to 0.630. Therefore, the \namchar score is able to classify many more Twitter users. The $t$-test shows that this improvement is significant, $t(9,059) = 3.27, p<.01$, in which the \namchar scores obtained better results (gender score: M=0.683, SD=0.277; \namchar score: M=0.692 SD=0.377). Therefore, concerning Research Question~\ref{q:namchar:score} we conclude that the performance of the gender score can indeed be improved by the use of \namchar.

\subsection{Gender Inference Methods}
\label{sec:methods}
In this section, we describe our modification of the Threshold Classifier of \citeauthor{liu2013whats}, including in particular the use of \namchar. We begin with recalling \citeauthor{liu2013whats}'s approach.

\subsubsection{Threshold Classifier}
\label{sec:threshold-classifier}
The Threshold Classifier of \citet{liu2013whats} uses a two-step approach. During the first step, the absolute value of the gender score is compared with a threshold value $\tau$ (set to 0.85 by \citet{liu2013whats}). The gender label that corresponds to this gender score is used directly if the absolute value is above the threshold, because these names are predominantly used by a single gender only.

In Step~2, the remaining user's names are classified by an SVM with a radial basis function\footnote{\citeauthor{liu2013whats} found the parameters for gamma and costs using a grid search technique, but did not mention them.} using a feature set that was developed in prior work \citep{liu2013whats,pennacchiotti2011machine,rustagi2009learning,burger2011discriminating}. These features consist of the following three groups:
\begin{itemize}
  \tightlist
  \item Features that are extracted from the full text messages (i.e., Tweets) of the given Twitter user. They consist of the $k$-top most discriminating words, $k$-top word stems\footnote{The stems are obtained by passing the words of all Tweets to the Lovins stemmer \citep{lovins1968development}.} (e.g., ``paper'' representing ``paper'', ``papers'', or ``papered''). $k$-top co-stems (\citet{lipka2011classifying} demonstrate that the stem-reduced words, or co-stems, yield a significant improvement over classical bag of words models; for example, ``s'' for ``papers''), $k$-top digrams (the most discriminating digrams; for example, ``pa'', ``ap'', ``pe'', and ``er''), $k$-top trigrams (the most discriminating trigrams; for example, ``pap'', ``ape'', and ``per''), and $k$-top hashtags (hashtags are labels that are attached by the users to their Tweet messages; for example, ``paper'' for ``\#paper'').
  \item Statistics that are extracted from the profile information of a given Twitter user: The average numbers of tweets, mentions, hashtags, links, and retweets per day, the ratio between retweets and tweets, and the friend-follower ratio.
  \item Gender information: The gender score of the name of the Twitter user as described in \secref{gnas} had been added as sole name-related feature by \citet{liu2013whats} in their original experiments.
\end{itemize}

The $k$-top full text features are computed by selecting a score for every term $t$ using $s(t) = |t_{\text{male}}| - |t_{\text{female}}|$, where $|t_{\text{\{gender\}}}|$ is the number of times the term was used by all users of the given gender. This results in one list of $k$-top terms for male and one for female; therefore, in $2 \times k$ features. The scores of the feature vector are then computed for every user $u$ using $s(u) = \frac{|t_u|}{|T_u|}$, where $t_u$ is the number of occurrences the term is used by the given user and $|T_u|$ is the number of all terms used by the same user.

\begin{table}[tb]
\centering
\caption{Examples for 3-top terms for each classification feature by gender.}
\label{tab:top-k-examples}
\begin{tabular}{lll}
  \hline
feature & male & female \\
  \hline
words & team, league, bro & hair, omg, cute \\
  stems & scor, win, team & girl, bab, feel \\
  co-stems & s, e, ed & n, i, a \\
  digrams & in, er, re & ee, ah, aa \\
  trigrams & ing, ion, ent & aha, eee, aaa \\
  hashtags & soundcloud, mufc, np & sorrynotsorry, excited, love \\
   \hline
\end{tabular}
\end{table}

\tabref{top-k-examples} shows as an example the 3-top terms for every feature category by gender that are obtained from all collected Twitter data. The complete list of 20-top terms for male names contains only verbs and conjunction words. The top words for female users on the other hand contain many more nouns. Men's hashtags tend to contain more technical or business tags, while women use more emotional tags.

\subsubsection{\namchar Threshold Classifier}
Our \namchar Threshold Classifier is a modification of \citeauthor{liu2013whats}'s Threshold Classifier. It varies in two aspects: First, we apply a pre-processing step to remove decorative elements as discussed in \secref{gnas}. Then, we use \namchar to assign a gender score to all names that are not found in the \census data.

\citeauthor{liu2013whats} discovered that they could not assign a meaningful gender score (i.e., $\neq0.0$) to 66~\% of the names they found on Twitter. Extending the name data by using the \namdict data instead of the \census data did not reduce the number of those instances significantly, as shown in \figref{gender-score-distributions}. Consequently, they would still lower the effect of the given name as a feature on the overall threshold classification. To this end, our goal is to further decrease the amount of those instances as much as possible by applying more pre-processing and adapting the way the gender score is computed.

One major challenge, while classifying the names of Twitter users is the multitude of special characters that are used in many user names. Therefore, we applied---unlike \citeauthor{liu2013whats}---the following pre-processing steps to all user names in the given order:

\begin{enumerate}
  \tightlist
  \item Latin to ASCII conversion: There are many regional letter variations of all vowels. We applied a Latin to ASCII conversion to all names in order to make the extraction of the number of vowels more robust (e.g., the vowel ``ü'' will be converted to the vowel ``u'').
  \item Removal of all non-letters: Next, we removed from the converted names all characters that are not letters or whitespaces. This step removes all numbers, punctuation signs, and emoji like flags and smilies. It enables us to classify decorated names (see \secref{gnas}).
\end{enumerate}

The pre-processed name strings are used to determine the \namchar gender score. To this end, the name string is split at white-spaces. Then we check if we can find any part in the currently applied name databases, going from left to right. This procedure ends as soon as we find our first match, in which case the gender score of the match is assigned to the name of interest. A score of $0.0$ is assigned, should no match be found. This reduces the number of unknown names on Twitter further by \preprocessingChange names or \preprocessingChangeShare~\%.

Next, we need a way to translate the predictions of \namchar to a meaningful gender score. It should reflect the reliability of the \namchar classifier. We use the probability of the classifier and linearly scale it to the gender score range of $-1.0$ (``female'') to $1.0$ (``male'').

Following \citeauthor{liu2013whats}, we used a value of 20 for the number $k$ of top terms and 0.85 for the threshold value $\tau$\footnote{Note that $\tau=0.85$ implies that ``mostly male/female name'', which obtained a gender score of 0.8 in \secref{gnas}, will be processed in Step~2 of the Threshold Classifier for further differentiation.}. We randomly split the dataset into halves and used the first partition to find the gamma and costs parameters for the SVM using a grid search technique. The best found parameters were then used to conduct the actual evaluation on the second partition.

In order to improve the gender classification of Twitter users, we evaluated both steps of the Threshold Classifier separately. First, we wanted to know whether or not it is possible to improve the performance for the first step where the threshold values decides the outcome. This part is independent from the actual classification and, therefore, cannot improve using our \namchar approach. Then, we tried to improve the classification performance on the remaining Twitter users using \namchar.

\begin{table}[tb]
\centering
\caption{Evaluation results for those data that where labeled in Step~1 by the threshold value showing the resulting performance values and the number of affected Twitter users~(N).}
\label{tab:svm-results-twitter-step-1}
\begin{tabular}{lrrrrr}
  \hline
Dataset & N & AUC & ACC & REC & K \\
  \hline
Census & 2962 & 0.912 & 0.912 & 0.922 & 0.824 \\
  nam\_dict & 2799 & 0.892 & 0.892 & 0.876 & 0.783 \\
   \hline
\end{tabular}
\end{table}

The findings on the gender score in \secref{gnas} indicated that the \census data is better suited than the \namdict data if applied on the whole dataset. But, only gender scores above the threshold value are relevant in Step~1. It is still possible that the \namdict data is better in generating such scores.  \tabref{svm-results-twitter-step-1} shows the performance that is achieved during this step. The data confirms that using the \census data results in better results in all four measures. Additionally, the gender score was able to label more Twitter users than with \namchar score.

The remaining 3,980 Twitter users were passed on to the classification step. 3,708 of those users have a gender score of 0.0, which leaves 272 users with a low gender score. Our expectation is that the classification of these user names---which cannot be done by a simple lookup in the database---will benefit more strongly from a more sophisticated classifier. To evaluate this assumption, we ran two experiments comparing the Threshold Classifier~(TH) with the \namchar Threshold Classifier~(TH + NC) on the \census data. Each experiment was conducted using a 10-fold cross validation with three repeats. The true gender was given by the labels assigned by the AMT workers in the ground truth data of \citeauthor{liu2013whats}, as described in \secref{data:twitter}.

\begin{table}[tb]
\centering
\caption{Evaluation results for those data that where labeled in Step~2 by the classifier showing the performance values and the gamma~(GAM) and costs~(C) parameter of the SVM. The methods column contains the user methods [i.e., Threshold Classifier~(TH) and NamChar Threshold Classifier~(TH + NC)].}
\label{tab:svm-results-twitter-step-2}
\begin{tabular}{lrrrrrr}
  \hline
Method & GAM & C & AUC & ACC & REC & K \\
  \hline
TH & 0.00400 & 1 & 0.892 & 0.814 & 0.813 & 0.628 \\
  TH + NC & 0.00379 & 1 & 0.896 & 0.820 & 0.815 & 0.639 \\
   \hline
\end{tabular}
\end{table}

\tabref{svm-results-twitter-step-2} shows the results for all names that have a gender score below threshold $\tau$. This is the part where the classification is executed using the full text, profile, and gender information. The use of \namchar results in a slight improvement compared to the plain Threshold Classifier. This is a further 4.4~\% reduction of the error rate compared to 0.892 of the Threshold Classifier without NamChar. The $t$-test shows that this difference between both results is significant, $t(3,979) = 2.04, p<.05$, in which the \namchar Threshold Classifier obtained better results  (TH: M=0.809, SD=0.151; NC + TH: M=0.811 SD=0.152). We used the classifiers probability $p$ for a certain class---using $p$ for male and $1 - p$ for female names---as input for the t-test. Therefore, concerning Research Question~\ref{q:namchar:overall}, we conclude that \citeauthor{liu2013whats}' approach can indeed be improved by assigning gender scores to those names that are not found in the database.

Overall, our experiments show that the influence of the user name data is more complex than originally expected. \citeauthor{liu2013whats} raised the concern that the \census data are not broad enough to be used on Twitter, because it contains only names that are used in the United States of America, whereas Twitter is used by people from all around the world. We could show, however, that a larger dataset did not necessarily improve the performance of the Threshold Classifier. The names that are not contained in the database are good candidates for improvement. We utilized them by assigning a score using the \namchar approach. Further,  \citeauthor{liu2013whats} were only able to assign scores to names from the US, while \namchar works in other countries and for ill-formed names as well.


\section{Conclusions and Future Work}
\label{sec:conclusion}
Our first contribution was the proposal of a novel name classifier---\namchar---that is able to label a given name as one of the two genders using only a small set of characteristics extracted from the written first name. We evaluated the performance of all possible name characteristics and selected the statistically significant characteristics.

Our second contribution was the introduction of an improved gender inference classifier for Twitter users. We observed that only using a larger name database will not improve the performance of the Threshold Classifier. The best results were achieved by our \namchar Threshold Classifier.

There are some issues that might be interesting to investigate further. It is for example surprising that \citeauthor{liu2012using} used only the Lovins stemmer to compute the $k$-top stems and co-stems during the feature computation. The Porter stemmer~\citep{porter1980algorithm} has become a de facto standard since then and has mostly replaced the Lovins stemmer. It would be interesting to compare the influence of both stemmers on the overall performance. We refrained from doing this here, because our intention was to study the effect of the gender score.

The further reduction of unknown names is also a field of further improvement. One step into this direction could be the discovery of personal names from nicknames. \citet{bollegala2011automatic} proposed a method for this task that could be tested on Twitter names.

Our ground truth data was already cleared from company accounts, which is not the case if we want to apply it on all Twitter users. Finding a reliable way to distinguish personal from company accounts would be required.

We considered some characteristics like the emphasized vowels or the finding of \citet{sidhu2015whats} about the application of the Bouba/Kiki effect as problematic, because we had no information about the proper pronunciation of the users' names. They could be more effective if one had information about the actual pronunciation, for example, when the origin of the Twitter user could be identified.

One general drawback of the threshold classification approach is its strong dependency on the used language. Tweets that are not written in English will create no meaningful feature vector. This could be solved by adding more language independent features like ``frequency statistics'', ``retweeting tendency'', or ``neighborhood size:'' to the feature vector. For instance, one could use text-independent approaches like using profile image attributes \citep{alowibdi2013language} or by extracting user attributes from the tweeted images \citep{merler2015you}.


\section{Acknowledgments}
We thank Stephan Doerfel, Janina Krawitz, Folke Mitzlaff, and Christoph Scholz for comments that greatly improved the manuscript.

\setlength{\bibsep}{2pt}
\bibliography{gender-inference-references}
\bibliographystyle{abbrvnat}

\balancecolumns

\end{document}